\documentclass[11pt,a4paper]{article}
\usepackage[hyperref]{emnlp-ijcnlp-2019}

\usepackage{times}
\usepackage{latexsym}
\usepackage{amsmath}
\usepackage{amssymb}
\usepackage{graphicx}
\usepackage[utf8x]{inputenc}
\usepackage{url}
\usepackage{multirow}
\usepackage{enumitem}

\aclfinalcopy 

\title{Low-Resource Sequence Labeling via Unsupervised Multilingual Contextualized Representations}

\author{Zuyi Bao$^\ddagger$,
	Rui Huang$^\diamondsuit$,
	Chen Li$^\ddagger$\thanks{\ \ Corresponding Author.} \and
	Kenny Q. Zhu\thanks{\ \ Kenny Q. Zhu was partially supported by NSFC grant 91646205 and Alibaba visiting scholar program.} \\
	$^\ddagger$Alibaba Group\\
	$^\diamondsuit$Zhejiang University, China\\
	$^\dagger$Shanghai Jiao Tong University, Shanghai, China 
	\\
	$^\ddagger$\texttt{\{zuyi.bzy,puji.lc\}@alibaba-inc.com} \\
	$^\diamondsuit$\texttt{iurgnauh@zju.edu.cn}, $^\dagger$\texttt{kzhu@cs.sjtu.edu.cn}
}
\date{}

\begin{document}
	\maketitle
	\begin{abstract}
		Previous work on cross-lingual sequence labeling tasks either requires parallel data or bridges the two languages through word-by-word matching. Such requirements and assumptions are infeasible for most languages, especially for languages with large linguistic distances, e.g., English and Chinese.
		In this work, we propose a Multilingual Language Model with deep semantic Alignment (MLMA) to generate language-independent representations for cross-lingual sequence labeling. 
		Our methods require only monolingual corpora with no bilingual resources at all and take advantage of deep contextualized representations. Experimental results show that our approach achieves new state-of-the-art NER and POS performance across European languages, and is also effective on distant language pairs such as English and Chinese.\footnote{The code is released at \url{https://github.com/baozuyi/MLMA}.}
	\end{abstract}

	\section{Introduction}
	Sequence labeling tasks such as named entity recognition (NER) and part-of-speech (POS) tagging are fundamental problems in natural language processing (NLP). Recent sequence labeling models achieve state-of-the-art performance by combining both character-level and word-level information~\cite{chiu16named,ma16end,lample16neural}. However, these models heavily rely on large-scale annotated training data, which may not be available in most languages. Cross-lingual transfer learning is proposed to address the label scarcity problem by transferring annotations from high-resource languages (source languages) to low-resource languages (target languages). In this scenario, a major challenge is how to bridge interlingual gaps with modest resource requirements.
	
	
	There is a large body of work exploring cross-lingual transfer through language-independent features, such as morphological features and universal POS tags for cross-lingual NER~\cite{tsai16cross} and dependency parsers~\cite{McDonald11multi}. However, these approaches require linguistic knowledge for language-independent feature engineering, which is expensive in low-resource settings. Other work relies on bilingual resources to transfer knowledge from source languages to target languages. Parallel corpora are widely used to project annotations from the source to the target side~\cite{yarowsky2001inducing,ehrmann11building,kim12multilingual,wang14cross}. These methods could achieve strong performance with a large amount of bilingual data, which is scarce in low-resource settings. 
	
	Recent research leverages cross-lingual word embeddings (CLWEs) to establish inter-lingual connections and reduce the requirements of parallel data to a small lexicon or even no bilingual resource~\cite{ni17weakly,fang2017model,xie2018neural}. However, word embedding spaces may not be completely isomorphic due to language-specific linguistic properties, and therefore cannot be perfectly aligned. For example, different from English, Chinese nouns do not distinguish singular and plural forms, while Spanish nouns distinguish masculine and feminine.
	
	On the other hand, NER tags such as person names, organizations, and locations are shared across different languages. Language-independent frameworks such as universal conceptual cognitive annotation~\cite{abend2013universal}, universal POS~\cite{petrov2011universal}, and universal dependencies~\cite{joakim2016universal} are defined to represent different languages in a unified formation. These work serves as our motivation to assume that the semantic meanings of words from different languages can be roughly aligned at a conceptual level and it is more reasonable to align deep semantic representations instead of shallow word embeddings. Meanwhile, monolingual contextualized embeddings derived from language models have shown to be effective for extracting semantic information and have achieved significant improvement on several NLP tasks~\cite{peters2018deep}.

	In this paper, we propose a Multilingual Language Model with deep semantic Alignment (MLMA). We train MLMA on monolingual corpora from each language and align its internal states across different languages. Then MLMA is utilized to generate language-independent representations and to bridge the gaps between high-resource and low-resource languages. For evaluation, we conduct extensive experiments on the NER and POS benchmark datasets under cross-lingual settings. The experiment results show that our methods achieve substantial improvements comparing to previous state-of-the-art methods in European languages. We also validate our approaches on a distant language pair, English-Chinese, and the results are competitive with previous methods which use large-scale parallel corpora. Our contributions are as follows:
	\begin{enumerate}\setlength{\itemsep}{-0.15cm}
		\item Instead of word-level alignment, we propose MLMA that uses contextualized representations to bridge the inter-lingual gaps.
		\item We propose three methods to align contextualized representations without any bilingual resource.
		\item Our methods achieve new state-of-the-art performance on cross-lingual NER and POS tasks in European languages, and very competitive results for English-Chinese NER, where previous work uses large parallel data.
	\end{enumerate}
	
	\section{Approach}
	Our approach belongs to the model transfer (Section~\ref{sec:model_transfer}) and mainly consists of three steps:
	\begin{enumerate}\setlength{\itemsep}{-0.1cm}
		\item Training a multilingual language model with alignment (MLMA) using monolingual corpora of the source and the target languages. (Section \ref{sec:lm_arch}, \ref{sec:lm_align} and \ref{sec:lm_train})
		\item Building a cross-lingual sequence labeling model based on the language-independent representations from the MLMA. (Section \ref{sec:clcr} and \ref{sec:seq_label})
		\item Learning the cross-lingual sequence labeling model (with MLMA fixed) on the annotated data of source languages and directly applying it to the target languages. 
	\end{enumerate}
	
\begin{figure*} 
		\centering
		\includegraphics[trim=0.5cm 5cm 1.3cm 3cm,clip,width=1.0\textwidth]{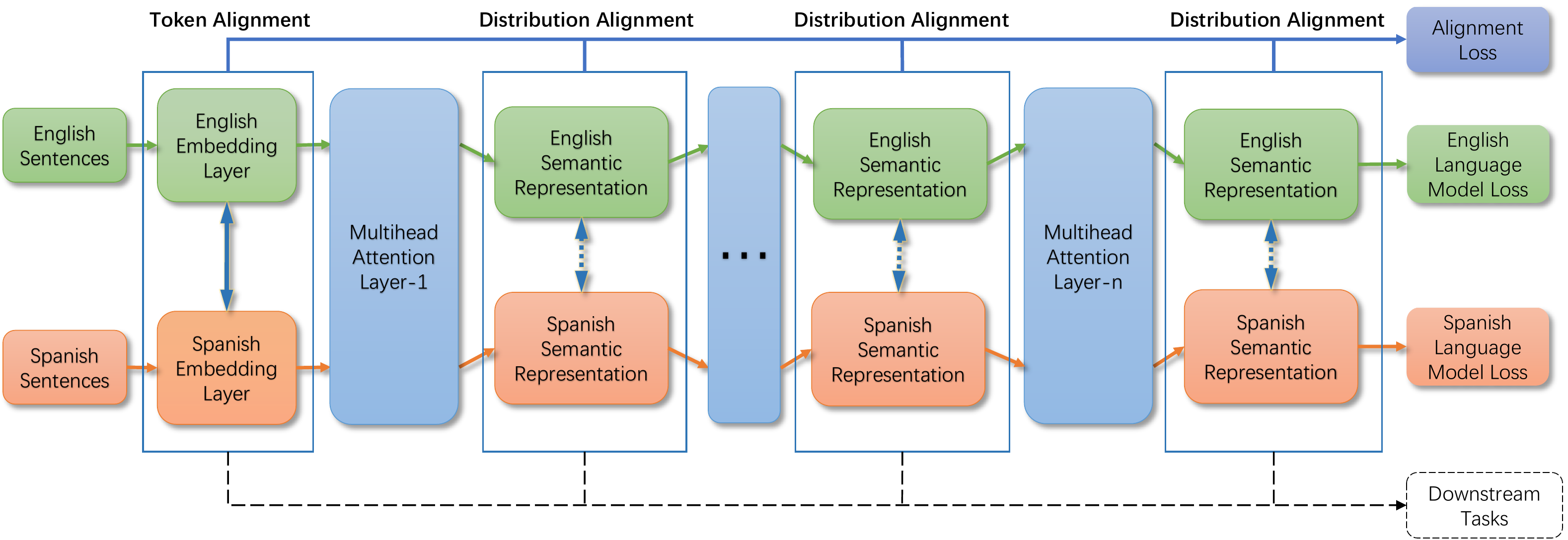}
		\caption{The architecture of MLMA consists of language-specific embedding layers and language-agnostic Transformer layers. MLMA is jointly learned through language modeling loss and alignment loss, and its internal representations are utilized to bridge the gap between source and target languages. }
		\vspace{-1.5em}
		\label{fig:main_architecture}
	\end{figure*}

	The architecture of MLMA is shown in Figure~\ref{fig:main_architecture}. In the following sections, we focus on introducing the Step 1 and 2.  We first present the architecture of MLMA and describe how to build the unsupervised multilingual alignment. Next, we propose effective methods for collapsing the multi-layer hidden states from MLMA into a single representation. Finally, we introduce the sequence labeling model used in the experiments.
	
	\subsection{Language Model Architecture} \label{sec:lm_arch}
	MLMA is a language model with multi-head self-attention mechanism~\cite{vaswani2017attention}. The architecture is similar to \newcite{radford2018improving}, except that we combine both a forward and a backward Transformer decoder to build a bidirectional language model. Take the forward direction as an example, given a sentence with $N$ tokens $W = [w_1, w_2, \cdots, w_N]^T$ as input, we first map the sequence of tokens $W$ to token embeddings $\overrightarrow{H}_0 \in \mathbb{R}^{N \times d} $:
	\setlength{\abovedisplayskip}{3pt}
	\setlength{\belowdisplayskip}{3pt}
	\begin{equation}
	\overrightarrow{H}_0 = WE_e + E_p
	\end{equation}
	where $E_e$ and $E_p$ are the embedding matrix and the positional encoding matrix, and $d$ is the dimension of embeddings and hidden states.
	
	Then $n$ blocks of transformer layers are stacked above the token embeddings. Each block contains a masked multi-head self-attention and a position-wise feedforward layer. The detailed implementation is the same as~\newcite{vaswani2017attention}.
	\begin{equation} \label{eq:transformer_layer}
	\overrightarrow{H}_l = {\rm TransformerLayer}(\overrightarrow{H}_{l-1})
	\end{equation}
	where $\overrightarrow{H}_l$ refers to the output of the $l$-th transformer block. Finally, the output distribution over the next tokens is calculated through a softmax function with tied embedding matrix.
	\begin{equation}
	\overrightarrow{P} = {\rm softmax}(\overrightarrow{H}_n E_e^T)
	\end{equation}
	For the backward direction, we calculate $\overleftarrow{H_l}$ and $\overleftarrow{P}$ in an analogous way. Finally, we jointly minimize the negative log likelihood of the forward and backward directions:
	\begin{equation} \label{eq:NLL}
	\begin{aligned} 
	NLL = - \sum_{t=1}^N (&\log p(w_t | w_1, \ldots, w_{t-1}) \\
	+ &\log p(w_t | w_{t+1}, \ldots, w_{N}))
	\end{aligned}
	\end{equation}
	
	In a multilingual setting, we share all parameters in Transformer layers across different languages to facilitate language-agnostic representations, except that we adopt an individual embedding matrix $E_e$ for each language.
	
	\subsection{Unsupervised Distribution Alignment} \label{sec:lm_align}
	
	We find that only sharing Transformer layers is not enough for forcing hidden representations from different languages into a common space, as suggested in experiments (Section~\ref{sec:result_ner}). Therefore, we propose three methods to build cross-lingual representations based on identical strings, mean/variance, and average linkage.
	
	To simplify the description, we take the alignment between two languages $s$ and $t$ as an example, but our methods can be directly extended to a scenario with multiple languages by adding the alignment between each pair of languages.
	
	\subsubsection*{Notation}
	For the language model, given a sentence with $N$ tokens, the forward internal representation  $\overrightarrow{H}_l$ in Eq~\eqref{eq:transformer_layer} can be expanded as $\overrightarrow{H}_l = [\overrightarrow{h}_{l,1}, \cdots, \overrightarrow{h}_{l,k}, \cdots, \overrightarrow{h}_{l,N}]^T$, where $\overrightarrow{h}_{l,k}$ refers to the forward hidden representation of the $k$-th token in the sentence. Then we concatenate the forward and backward hidden representations for each token, $h_{l,k} = \overrightarrow{h}_{l,k} \oplus \overleftarrow{h}_{l,k}$.
	
	We denote the collection of the token representations ${h}_{l,k}$ at layer $l$ from the whole corpora of language $s$ as $C_l^s$, which can be regarded as a sampling from the deep semantic space of language $s$. Similarly, $C_l^t$ is used for language $t$.
	
	\subsubsection*{Identical Strings}
	Similar language pairs such as English and Spanish have a large number of identical strings shared between their vocabularies, which are utilized as the seed dictionary for embedding alignment in previous work~\cite{smith2017offline}. Similarly, we treat identical strings as explicit supervision signals and align the embeddings of identical strings between different languages. The matching of the embeddings from different languages will lead to an implicit alignment of internal representations. In the experiments, we directly minimize the Euclidean distance between the embeddings of each identical string across different languages:
	\begin{equation*}
	L_{iden} = \dfrac{\lambda^{iden}}{| W_{iden}^{(s,t)} |} \sum_{w \in W_{iden}^{(s,t)}} ||e_{w}^s - e_{w}^t||
	\end{equation*}
	where $W_{iden}^{(s,t)}$ is the set of identical strings between the vocabulary of language $s$ and language $t$, and $|W_{iden}^{(s,t)}|$ refers to the number of members in $W_{iden}^{(s,t)}$. $\lambda^{id}$ is a scaling weight, and $e_{w}^s$ ($e_{w}^t$) is the embedding of word $w$ from embedding matrix $E_e^s$ ($E_e^t$) of language $s$ ($t$). 
	
	
	
	\subsubsection*{Mean and Variance}
	In this section, we propose another approach to directly align the distributions of internal representations between different languages. In particular, we leverage the mean and variance of internal distributions for alignment. We denote the mean and variance of $C_l^s$ as $m_{l}^s$ and $v_{l}^s$. Similarly, $m_{l}^t$ and $v_{l}^t$ refer to the mean and variance of $C_l^t$. We minimize the Euclidean distance between the mean and variance of language $s$ and language $t$ for all layers:
	\begin{equation*}
	L_{mv} = \sum_{l=0}^n (\lambda_{l}^m \cdot \dfrac{||m_{l}^s - m_{l}^t||}{|m_{l}^s| + |m_{l}^t|} + \lambda_{l}^v \cdot \dfrac{||v_{l}^s - v_{l}^t||}{|v_{l}^s| + |v_{l}^t|})
	\end{equation*}
	where $\lambda_{l}$ is a scaling weight, and $|\cdot|$ is the L1 norm of a vector. Without the denominators, the model could escape this regularization by learning a mean and variance with low absolute values.
	
	In practice, rather than calculating the mean and variance over the whole source and target corpora, we use the mean and variance of the source and target inner states ${h}_{l,k}$ in the current mini-batch as an approximation.
	
	\subsubsection*{Average Linkage}
	In this method, we employ another metric, average linkage, to perform a more precise point-wise matching. The average linkage is a widely used metric for calculating the similarity of clusters and networks~\cite{yim2015hierarchical,seifoddini1989single,newman2012communities,moseley2017approximation}. It is sensitive to the shape, thus serves as a better choice than mean and variance. The average linkage measures the similarity of two sets $X$ and $Y$ by calculating the averaged distance between all members of each set:
	\begin{equation*}
	avl(X, Y) = \dfrac{1}{n_{X} \cdot n_{Y}} \sum_{x \in X} \sum_{y \in Y} f(x, y)
	\end{equation*}
	where $n_X$ ($n_Y$) is the number of members in $X$ ($Y$), and $f$ is a distance function. We take Euclidean distance as the distance function $f$ and minimize the average linkage between $C_l^s$ and $C_l^t$:
	\begin{equation*}
	\begin{aligned}
	L_{avl} = \sum_{l=0}^n & \lambda_{l}^{avl} \cdot [2 \cdot avl(C_l^s, C_l^t) \\
	& - avl(C_l^s, C_l^s)  - avl(C_l^t, C_l^t)]
	\end{aligned}
	\end{equation*}
	Similarly, the terms $avl(C_l^s, C_l^s)$ and $avl(C_l^t, C_l^t)$ are used to prevent the model from escaping this regularization. In practice, we calculate $L_{avl}$ between the source and target inner states ${h}_{l,k}$ inside the mini-batch as an approximation.
	
	The regularization $L_{avl}$ is similar to the maximum mean discrepancy (MMD), which is often employed in domain adaptation~\cite{tzeng2014deep,long2015learning} and style transfer~\cite{li2017demystifying} for images. However, different from MMD, our method directly uses Euclidean distance instead of the kernel function.
	
	\subsection{Training of MLMA} \label{sec:lm_train}
	During the training stage of MLMA, we sample equivalent number of sentences from the monolingual corpora of each language for each mini-batch. Then MLMA is optimized through a combination of the language modeling loss $L_{lm}$ and the alignment regularization loss $L_{reg}$. For each alignment method, we use its corresponding alignment loss:
	\begin{equation*}
	\begin{aligned}
	L = &L_{lm} + L_{reg} \\
	{\rm where}\ \ &L_{lm} =  \sum_{i \in \{s, t\}} \lambda_{i}^{lm} \cdot NLL_i, \\
	&L_{reg} \in \{L_{id}, L_{mv}, L_{avl}\}
	\end{aligned}
	\end{equation*}
	where $\lambda_{i}^{lm}$ is used for balancing the convergence speed of different languages. $NLL_i$ is the negative log likelihood of language $i$ in Eq~\eqref{eq:NLL}.
	
	\subsection{Cross-lingual Representations} \label{sec:clcr}
	After the MLMA is trained, we fix its parameters and extract the hidden states as cross-lingual contextualized representations (CLCRs). In this section, we propose two effective strategies for integrating these multi-layer high-dimensional representations into downstream models.
	
	\noindent\textbf{Self-Weighted Sum} For each token, we concatenate all layers of hidden states and feed them into a multi-layer perceptron (MLP) to calculate a $(n+1)$-dimensional weight vector, $s = {\rm softmax}({\rm MLP}(h_{0,k} \oplus \cdots \oplus h_{n,k}))$. Then we calculate a weighted sum of these layers according to the weight vector, $\textbf{CLCR}_k = \sum_{l=0}^n s_{l}\cdot h_{l,k}$.
	
	\noindent\textbf{Fully-Weighted Sum} We introduce a weight matrix, $F \in \mathbb{R}^{(n+1) \times 2d}$, with separate weights for each hidden dimension. The weight matrix $F$ is softmaxed by column and used to calculate a weighted sum of all layers for each hidden dimension, $\textbf{CLCR}_k = \sum_{l=0}^n F_l \odot h_{l,k}$, where $\odot$ is the element-wise product.
	
	The parameters of the ${\rm MLP}$ and $F$ are trained during the learning of sequence labeling model.
	
	
	\subsection{Sequence Labeling Model} \label{sec:seq_label}
	The sequence labeling model is then built on the CLCRs. For both NER and POS tasks, we use an LSTM-CRF model following~\newcite{lample16neural}, which consists of a character-level LSTM, a word-level LSTM, and a linear-chain CRF. 
	
	More specifically, given a sequence of words as $[w_1, w_2, \ldots, w_N]$, where $w_k$ is composed of a sequence of characters $[c_{k,1}, c_{k,2}, \ldots, c_{k,m}]$. First, for each word $w_k$, the character-level LSTM takes its character sequence $[c_{k,1}, c_{k,2}, \ldots, c_{k,m}]$ as input and outputs a vector $e_k$ to represent this word. Then the pre-trained $\textbf{CLCR}_k$ is concatenated with $e_k$ to form a word-level embedding $x_k$. Finally, the sequence of word-level embeddings $[x_1, x_2, \ldots, x_N]$ are fed into the word-level LSTM, and the linear-chain CRF are employed to predict the probability distribution for all possible output label sequences.
	
	
	\section{Experiments}
	
	We first introduce the datasets used in the experiment and then the implementation details of our models, before presenting the results on NER and POS tasks.
	
	\subsection{Datasets} 
	For cross-lingual NER, we evaluate the proposed approaches on CoNLL 2002/2003 datasets~\cite{tksintro2002conll,tjongkimsang2003conll}, which contain four European languages, English (en), Spanish (es), Dutch (nl), German (de) and four entity types (person, location, organization, and MISC). We also evaluate a distant language pair, English-Chinese, on OntoNotes(v4.0) dataset~\cite{hovy2006ontonotes}. We adopt the same dataset split and four valid entity types (person, location, organization, and GPE) as described in \cite{wang14cross}.
	
	For cross-lingual POS, we use the Danish (da), Dutch (nl), German (de), Greek (el), Italian (it), Portuguese (pt), Spanish (es) and Swedish (sv) portion from CoNLL 2006/2007 dataset~\cite{buchholz2006conll,nivre07the}. Following previous work \cite{fang2017model}, we train the sequence labeling model on Penn Treebank data and adopt the universal POS tagset~\cite{petrov11auniversal}.
	
	In all cases, the sequence labeling model is trained on the source language (English) training data and is tested on the target language test data.
	
	
	\subsection{Details of MLMA}  
	We adopt a 6-layer bi-directional Transformer decoder with 8 attention heads. The dimension size of hidden states and inner states are 512 and 2048, respectively. The dropout rates after attention and residual connection are both 0.1. We use the Adam optimization scheme~\cite{kingma2014adam} with a learning rate of 0.0001 and a gradient clip norm of 5.0. The vocabulary size of each language is 200,000, and we train the model with a sampled softmax~\cite{jean15on} of 8192 samples. We only keep the sentences containing less than 200 tokens for training and group them into batches by length. Each batch contains around 4096 tokens for each language. The language modeling weight $\lambda_i^{lm}$ is set to be 1.0 for each language. For alignment, $\lambda_l^m$, $\lambda_l^v$, $\lambda_l^{al}$ are set to be 0.1, 0.01 and 1.0 for every layer $l$, and $\lambda^{iden}$ is set to be 100.
	
	For languages except English, the latest dump of Wikipedia is used as monolingual corpora. For English, we use 1B Word Benchmark~\cite{chelba2013one} to reduce the effects of potential internal alignment in Wikipedia~\cite{zirikly15cross,tsai16cross}. 
	
	All characters are preprocessed to lowercase, and Chinese text are converted into the simplified version through OpenCC\footnote{\url{https://github.com/BYVoid/OpenCC}}. The corpora of European languages are tokenized by nltk~\cite{loper2002nltk} and Chinese text is segmented using Ltp\footnote{\url{https://github.com/HIT-SCIR/pyltp}}.
	
	\subsection{Details of Sequence Labeling Model} 
	In our experiments, we set the hidden size of word-level LSTM and character-level LSTM to be 300 and 100, respectively. The character embedding size is set to be 100. We apply dropout at both the input and the output of word-level LSTM to prevent overfitting. The dropout rate is set to be 0.5. We train the sequence labeling model for 20 epochs using Adam optimizer with a batch size of 20 and perform an early stopping when there is no improvement for 3 epochs. We set the initial learning rate to be 0.001 and decay the learning rate by 0.1 for each epoch. We do not update the pre-trained cross-lingual deep representations from MLMA during training. For each model, we run it five times and report the mean and standard deviation. We disable the character-level LSTM in English-German and English-Chinese NER as they have a different character pattern from English. For POS, we disable the character-level LSTM following~\newcite{fang2017model}.
	
	

	\subsection{Results for NER} \label{sec:result_ner}
	\begin{table*}[th]
		\small
		\centering
		\begin{tabular}{l c c c l} 
			Model  &  es & nl & de & Extra Resources\\
			\hline
			MLM (w/o alignment) + s.w.s. & 21.16 ± 1.40 & 33.97 ± 1.49 & 15.46 ± 1.21 & None\\
			MLM (w/o alignment) + f.w.s. & 23.61 ± 2.33 & 32.94 ± 1.62 & 16.38 ± 1.09 & None\\
			\hline
			\newcite{tackstrom12cross} & 59.30 & 58.40 & 40.40 & parallel corpus\\
			\newcite{nothman13learning} & 61.00 & 64.00 & 55.80 & Wikipedia\\
			\newcite{wang14cross} & - & - & 60.00 & parallel corpus\\
			\newcite{tsai16cross} & 60.55 & 61.60 & 48.10 & Wikipedia\\
			\newcite{ni17weakly} & 65.10 & 65.40 & 58.50 & Wikipedia, parallel corpus, 5K dict.\\
			\newcite{mayhew17cheap} & 65.95 & 66.50 & 59.11 & Wikipedia, 1M dict.\\
			\newcite{xie2018neural} & 72.37 & 71.25 & 57.76 & None\\
			MUSE* & 66.17 ± 1.15 & 65.52 ± 0.78 & 55.46 ± 0.59 & None\\
			Multilingual BERT* & 66.42 ± 1.15 & 69.21 ± 0.48 & \textbf{70.78} ± 0.36 & None\\
			\hline
			MLMA-Iden + s.w.s. & 69.45 ± 0.91 & 68.82 ± 0.82 & 55.75 ± 1.64 & None\\
			MLMA-Iden + f.w.s. & 67.10 ± 0.78 & 68.15 ± 0.67 & 55.25 ± 1.29 & None\\
			MLMA-Mv + s.w.s. & 73.81 ± 0.83 & 70.61 ± 1.79 & 57.70 ± 0.71 &  None\\
			MLMA-Mv + f.w.s. & 74.12 ± 1.00 & 71.72 ± 0.70 & 57.84 ± 0.80 &  None\\
			MLMA-Avl + s.w.s. & 75.01 ± 0.79 & 76.22 ± 0.42 & 60.98 ± 1.00 & None \\
			MLMA-Avl + f.w.s. & 74.43 ± 0.50 & 76.02 ± 0.55 & 60.50 ± 0.43 & None \\
			MLMA-Avl (init) + s.w.s. & 75.72 ± 0.80 & \textbf{76.90} ± 0.30 & 63.01 ± 0.83 & None\\
			MLMA-Avl (init) + f.w.s. & 76.30 ± 0.76 & 76.85 ± 0.43 & 62.85 ± 0.47 & None\\
			MLMA-Avl (multi) + s.w.s. & \textbf{79.36} ± 0.57 & 74.89 ± 0.28 & 65.93 ± 0.32 & None\\
			MLMA-Avl (multi) + f.w.s. & 79.34 ± 0.35 & 74.74 ± 0.40 & 66.53 ± 0.35 & None\\
			\hline
		\end{tabular}
		\caption{ NER F1 scores on test sets of European languages. For previous work which reports multiple results, we only list their best performance on each language. Results of methods with mark * are obtained by running their released source code or models. The results of MUSE embeddings are produced by using them for direct model transfer. ``MLM" denotes our multilingual language model without alignment. The three alignment methods, \textbf{Iden} =  identical strings, \textbf{Mv} = mean and variance, \textbf{Avl} = average linkage, respectively. ``s.w.s" and ``f.w.s." are self-weighted sum and fully-weighted sum. ``init" represents using MUSE to initialize the embedding matrices in the MLMA. ``multi" refer to the multi-source transfer.}
		\label{table:NERmain}
	\end{table*}

	\begin{table}[t]
		\small
		\centering
		\begin{tabular}{l c} 
			Model  & zh \\
			\hline
			\newcite{wang14cross}$\diamond$ & \textbf{64.40} \\
			MUSE* & 35.35 ± 0.84 \\
			\newcite{xie2018neural}* & 44.13 ± 1.49 \\
			\hline
			\textit{Our methods} & \\
			MLMA-Iden + s.w.s. & 11.08 ± 0.89 \\
			MLMA-Iden + f.w.s. & 11.17 ± 0.69 \\
			MLMA-Avl + s.w.s. & 50.11 ± 1.51 \\
			MLMA-Avl + f.w.s. & 45.88 ± 2.49 \\
			MLMA-Avl (init) + s.w.s & 60.33 ± 1.39 \\
			MLMA-Avl (init) + f.w.s & 58.92 ± 1.22 \\
			\hline
		\end{tabular}
		\caption{ NER F1 scores on test sets for Chinese. The notations are the same as Table~\ref{table:NERmain}. Methods with mark $\diamond$ require parallel corpora.}
		\label{table:NERchs}
	\end{table}

	\begin{table*}[th]
		\centering
		\begin{tabular}{l c c c c c c c c c c}
			Model  & es & nl & de & da & el & it & pt & sv & Avg. \\
			\hline
			\newcite{das2011unsupervised}$\diamond$ & \textbf{84.2} & 79.5 & 82.8 & 83.2 & \textbf{82.5} & 86.8 & \textbf{87.9} & 80.5 & 83.31\\
			\newcite{fang2017model}$\dagger$ & 68.40 & 64.50 & 65.90 & 73.5 & 65.5 & 64.8 & 67.8 & 66.0 & 67.05 \\
			\newcite{fang2017model}$\dagger\ddagger$ & 81.20 & 82.30 & 78.90 & 81.9 & 80.1 & 81.9 & 82.1 & 78.1 & 80.81\\
			\newcite{xie2018neural}* & 73.25 & 75.46 & 80.72 & 29.75 & 71.65 & 71.19 & 76.48 & 64.36 &  67.86 \\ 
			MUSE* & 78.30  & 80.84  & 81.10  & 73.99  & 63.16  & 80.63  & 82.79  & 66.38 & 75.90\\
			Multilingual BERT*  & 83.86  & 84.79   & \textbf{87.16} & \textbf{83.77}  & 82.27 & \textbf{88.39}  & 87.86 &  \textbf{81.07} & \textbf{84.90} \\ 
			\hline
			\textit{Our methods} & & & & & & & &\\
			MLMA-Avl + s.w.s. & 81.60  & 85.10  & 84.10  & 83.38  & 77.04  & 84.46  & 86.93  & 80.78 & 82.92 \\
			MLMA-Avl + f.w.s. & 81.20  & 85.54  & 84.92  & 83.45  & 77.48  & 84.80  & 87.43  & 80.72 & 83.19 \\
			MLMA-Avl (init) + s.w.s. & 82.73  & 85.79  & 85.76  & 82.44  & 80.55  & 86.76  & 85.99  & 79.75 & 83.72 \\
			MLMA-Avl (init) + f.w.s. & 82.27  & \textbf{85.97}  & 86.37  & 82.25  & 81.31  & 86.28   & 86.53  & 80.00 & 83.87 \\
			\hline
		\end{tabular}
		\caption{ POS accuracy on test sets of European languages. The notations are consistent with Table~\ref{table:NERmain}. \newcite{fang2017model} report different results according to different resource requirements. We only list their best results in each setting. Methods with mark $\diamond$, $\dagger$, $\ddagger$ require parallel corpora, bilingual lexicons, and training data respectively.}
		\label{table:POSmain}
	\end{table*}
	
	We first train a multilingual language model without alignment (MLM) and report its performance of cross-lingual NER in Table~\ref{table:NERmain}. The poor performance demonstrates that only sharing part of the parameters in a language model is far from enough for cross-lingual transfer.
	
	As shown in Table~\ref{table:NERmain}, the mean/variance alignment strategy (MLMA-Mv) is competitive with previous work which utilizes extra bilingual resources (Section \ref{sec:annotation_projection}). The average linkage strategy (MLMA-Avl) performs a more precise alignment and gains a further improvement. We conducted experiments of using all three alignments together, and results show no significant improvement over average linkage alone. These results agree with our statements that average linkage performs a more precise matching, and thus, carries the benefits brought by the other methods.
	
	To demonstrate the strengths of the proposed cross-lingual contextualized representations (CLCRs) over cross-lingual word embeddings (CLWEs), we also report the results of using CLWEs for direct model transfer in Table~\ref{table:NERmain}.  Specifically, we compare with the unsupervised method MUSE from~\newcite{conneau2017word}. The experiment results demonstrate its effectiveness for cross-lingual sequence labeling. The alignment method using identical strings (MLMA-Iden) outperforms MUSE, suggesting that the contextual-level representations are more effective than the word-level ones. The other proposed methods (MLMA-Mv and MLMA-Avl) achieve significant improvement over MUSE and MLMA-Iden, which shows the benefit of directly aligning the contextualized representations.
	
	\noindent\textbf{Combination with CLWEs} We further demonstrate that CLWEs are compatible with our methods by using MUSE embeddings to initialize the embedding layer of our multilingual language model. The results of MLMA-Avl (init) shown in Table~\ref{table:NERmain} indicate that the CLWEs lead to a better initialization and improved performance.
	
	\noindent\textbf{Multi-source Transfer} We conduct experiments of multi-source transfer based on method MLMA-Avl and report the performance as MLMA-Avl (multi) in Table~\ref{table:NERmain}. The experiment settings largely follow the previous work~\cite{mayhew17cheap}. They employ two source languages for each target language and use syntactic features to choose the related source languages. For Spanish and German, we use English and Dutch as source languages. English and Spanish are adopted for Dutch. The multi-source transfer leads to a significant improvement for Spanish and German, but a slight decline for Dutch. In the follow-up experiment, we find that the Spanish training set achieves a poor cross-lingual performance on Dutch. Similar results are observed in the experiments of Spanish to English and Dutch to English. These results suggest that the cross-lingual transfer may be directional, and we leave this issue for future work.
	
	\noindent\textbf{Comparison with BERT} We also compare the performance of our MLMA with the released multilingual BERT~\cite{devlin2018bert}. As shown in Table~\ref{table:NERmain}, our MLMA-Avl achieves a better performance on Spanish and Dutch. For German, BERT achieves a high performance as it employs effective subword information through BPE. The architecture of BERT also performs better than LSTM.
	
	It is worth mentioning that, in previous work and this work, the corpora used in the experiments are limited to the source and the target language. In contrast, the multilingual BERT is jointly learned on Wikipedia of 102 languages and may benefit from a multi-hop transfer. BERT employs a shared BPE vocabulary for different languages, which implicitly performs a subword alignment similar to MLMA-Iden. Meanwhile, the proposed MLMA-Mv and MLMA-Avl methods are compatible with BERT and can be used to align the inner states of BERT.
	
	
	\begin{table*}[th]
		\small
		\centering
		\begin{tabular}{p{4.2cm}|p{10.5cm}} 
			\hline
			\multicolumn{2}{l}{\textbf{MUSE}} \\
			\hline \hline
			\multicolumn{2}{l}{ brown:\ oliváceo (olive), negruzcas (blackish), negruzco (blackish), marrón (brown), ocráceo (ochraceous) } \\ 
			\multicolumn{2}{l}{ chair:\ vicepresidenta (vice president), vicedecano (vice dean), cátedra (chair), vicedecana (vice dean), catedrático (professor) } \\ 
			\hline \hline
			\multicolumn{2}{l}{}\\
			\hline
			\multicolumn{2}{l}{\textbf{MLMA-Avl}} \\
			\hline \hline
			\multirow{2}{*}{\shortstack[l]{\textbf{[Brown]}'s office told news outlets \\ of his visit to Afghanistan ...}} & \textbf{[Neira]} escapó meses después rumbo a Miami para ... \\
			& (\textbf{Neira} escaped months later heading to Miami to ...) \\ \hline
			\multirow{2}{*}{\shortstack[l]{Wearing a \textbf{[brown]} suit with \\ matching hat, ...}} & La corona y vientre del macho son de un \textbf{[verde]} esmeralda brillante iridiscente, ...\\
			& (The crown and belly of the male are of an iridescent bright emerald \textbf{green}, ...) \\
			\hline \hline
			\multirow{2}{*}{\shortstack[l]{Sweden currently holds \\ the EU \textbf{[chair]}.}} & Tras tomar posesión de su \textbf{[asiento]} , Lois decide limpiar el lago para empezar, ... \\
			& (After taking possession of her \textbf{seat}, Lois decides to clean the lake to begin, ... )\\ \hline
			\multirow{2}{*}{\shortstack[l]{It's an honor to be asked to \\ \textbf{[chair]} the Man Booker Prize, ...}} &  ..., y fue la primera mujer en \textbf{[presidir]} un sindicato AFL-CIO. \\
			& (..., and was the first woman to \textbf{preside} over an AFL-CIO union.) \\
			\hline \hline
		\end{tabular}
		\caption{English words and their nearest Spanish words according to MUSE and MLMA-Avl.}
		\label{table:example}
	\end{table*}

	\subsection{A Case Study of Chinese NER}
	We conduct experiments and evaluate our approaches on a distant language pair, English-Chinese. The experiment results are shown in Table~\ref{table:NERchs}. \newcite{wang14cross} utilize 80K parallel sentences for annotation projection and report a strong performance. As Chinese and English do not share the alphabet, the number of identical strings is significantly smaller than similar languages pairs such as English-Spanish. Therefore, the MLMA-Iden achieves a lower result comparing to MUSE which uses adversarial training. The MLMA-Avl method performs a direct alignment of internal representations and achieves a significant improvement over the word-level methods. The initialization from CLWEs also proves its effectiveness for distant language pairs by gaining further improvement and reaching a comparable result with \newcite{wang14cross}. This experiment suggests that cross-lingual transfer is still challenging between distant language pairs.
	
	
	
	\subsection{Results for POS}
	We evaluate our methods on another sequence labeling task POS, and the results are shown in Table~\ref{table:POSmain}. We compare with previous studies using unsupervised cross-lingual clustering~\cite{fang2017model} and large-scale parallel corpora~\cite{das2011unsupervised}. As shown in Table~\ref{table:POSmain}, our models with deep semantic alignment outperform previous lexicon-based cross-lingual clustering by a large margin. When comparing to the previous method with a small amount of training data, the MLMA-Avl method obtains an improved accuracy without training data in the target languages. For further comparison, We also list the performance of applying the method from \newcite{xie2018neural} and multilingual BERT to the POS task.\footnote{The poor performance of \newcite{xie2018neural} on en-da is due to the low quality of word translation pairs generated by their method.}
	
	POS mainly relies on the information of each single word, and parallel corpora providing word alignment are effective for cross-lingual POS. Thus, previous annotation projection methods through parallel corpora are strong approaches for cross-lingual POS and often achieve a significantly better performance against previous unsupervised methods. The experimental results show that the proposed CLCRs are competitive and even achieve better average accuracy.
	
	\subsection{Self-Weighted v.s. Fully-Weighted Sum}
	As shown in Table~\ref{table:NERmain}, \ref{table:NERchs} and \ref{table:POSmain}, we observe that Self-Weighted Sum (SWS) generally outperforms Fully-Weighted Sum (FWS) in NER tasks, while the opposite is true for POS tasks. SWS allows weights to vary at each position in a sequence, while FWS imposes adaptive weights for each hidden dimension. We hypothesize that NER is more context-sensitive and requires models to adapt to different context information, which makes SWS a better option. On the other hand, the POS of words is more independent across different context, but certain feature dimensions in contextualized representations may be critical for making a judgment. Therefore, FWS has the edge over SWS for its ability to select out these dimensions.
	
	\section{What is Connected during Alignment?}
	In this section, we dive into the MLMA and investigate the question of what is connected between different languages during the alignment. From English 1B and Spanish Wikipedia, we randomly select 1,000 sentences for each language and extract their cross-lingual contextual representations using our MLMA-Avl model. We calculate the nearest neighbors in cosine distance for each word, and some of them are listed in Table~\ref{table:example}.
	
	In these cases, the MLMA can disambiguate word senses according to context information. For example, for the word \textit{brown} in English, the MLMA groups color \textit{brown} with \textit{verde} (green), and name \textit{Brown} with \textit{Neira} (a person name in Spanish) in the Spanish corpus. The proposed method is different from unsupervised translation in that, instead of learning a precise matching between English and Spanish words, the CLCRs establishes a high-level semantic connection between the source and the target language. The next example demonstrates that the MLMA is able to distinguish the part-of-speech of words. It connects an English verb \textit{chair} with a Spanish verb \textit{presidir} (preside), while a noun \textit{chair} with a noun \textit{asiento} (seat) in Spanish. To compare with unsupervised cross-lingual word embeddings, we list the top 5 similar words calculated using MUSE. As shown in Table~\ref{table:example}, MUSE successfully groups the English word \textit{brown} with Spanish words that are related to colors. However, without the help of contextual information, its ability of word sense disambiguation is limited. 
	
	\section{Related Work}
	\label{sec:related}
	Previous work in cross-lingual transfer learning can be roughly divided into two main branches: annotation projection and model transfer.
	
	\subsection{Annotation Projection}
	\label{sec:annotation_projection}
	In annotation projection approaches, parallel or comparable corpora are commonly used \cite{yarowsky2001inducing,ehrmann11building,das2011unsupervised,li12joint,tackstrom2013token,wang14cross,ni17weakly}. The source language sentences of parallel corpora are first annotated either manually or by a pre-trained tagger. Then, annotations on the source side are projected to the target side through word alignment to generate distantly supervised training data. Finally, a model of the target language is trained on the generated data. Wikipedia contains multilingual articles for various topics and can thus be used to generate parallel/comparable corpora or even weakly annotated target language sentences~\cite{kim12multilingual}.
	
	However, parallel corpora and Wikipedia can be rare for true low-resource languages. \newcite{mayhew17cheap} reduce the resource requirement by proposing a cheap translation method, which ``translates" the training data from the source to the target language word by word through a bilingual lexicon. While \newcite{xie2018neural} reduce the requirement of bilingual lexicons by an unsupervised word-by-word translation through CLWEs.
	
	\subsection{Model Transfer}
	\label{sec:model_transfer}
	Model transfer methods train a model on the source language with language-independent features. Thus, the trained model can be directly applied to the target language.
	
	\newcite{McDonald11multi} design a cross-lingual parser based on delexicalized features like universal POS tags. \newcite{tackstrom12cross} reveal that cross-lingual word cluster features induced using large parallel corpora are useful. Lexicon and Wikipedia also demonstrate effectiveness for language-independent feature engineering.  \newcite{zirikly15cross} generate multilingual gazetteers from the source language gazetteers and comparable corpus. Page categories and linkage information to entries from Wikipedia are extracted as strong language-independent features (wikifier features)~\cite{tsai16cross}. \newcite{bharadwaj16phonologically} facilitate the cross-lingual transfer through phonetic features, which work well between languages like Turkish, Uzbek, and Uyghur, but are not strictly language independent. Recently, CLWEs are used as language-invariant representations for direct model transfer in NER~\cite{ni17weakly} and POS~\cite{fang2017model}.
	
	Some of the previous work also proposes sequence labeling models with shared parameters between languages for performing cross-lingual knowledge transfer~\cite{lin18multi,cotterell17low,yang2017transfer,ammar16many,kim17cross}. However, these models are usually obtained through joint learning and require annotated data from the target language.
	
	\section{Conclusion}
	In this paper, we focused on a low-resources cross-lingual setting and proposed transfer learning methods based on the alignment of deep semantic spaces between different languages. The proposed multilingual language model bridges different languages by automatically learning cross-lingual disambiguated representations. Abundant NER and POS experiments are conducted on the benchmark datasets. Experimental results show that our approaches using only monolingual corpora achieve improved performance comparing to previous strong cross-lingual studies with extra resources.
	
	\bibliography{acl2019}
	\bibliographystyle{acl_natbib}
\end{document}